\documentclass{article} % For LaTeX2e
\usepackage{iclr2020_conference,times}
\usepackage{amsmath,amsfonts,bm}

\def\eqref#1{equation~\ref{#1}}
\def\1{\bm{1}}

\DeclareMathAlphabet{\mathsfit}{\encodingdefault}{\sfdefault}{m}{sl}
\SetMathAlphabet{\mathsfit}{bold}{\encodingdefault}{\sfdefault}{bx}{n}

\usepackage{hyperref}
\usepackage{url}
\usepackage{float}
\usepackage{xcolor}
\usepackage[noend,ruled,linesnumbered]{algorithm2e} % linesnumbered
\usepackage{multirow}

\newcommand{\toolname}[0]{PUPPER}

\title{Prioritized Unit Propagation with Periodic Resetting is (Almost) All You Need for Random SAT Solving}
\author{Xujie Si$^{1\dagger}$\thanks{Work done as an intern at DeepMind.}, Yujia Li$^{2\dagger}$, Vinod Nair$^{2\dagger}$, Felix Gimeno$^{2}$\\
$^{1}$University of Pennsylvania, $^{2}$DeepMind\\
$^\dagger$\text{Equal contributions}\\
\texttt{xsi@cis.upenn.edu, \{yujiali,vinair,fgimeno\}@google.com} \\
}

\iclrfinalcopy % Uncomment for camera-ready version, but NOT for submission.
\begin{document}

\maketitle

\begin{abstract}
We propose \emph{prioritized unit propagation with periodic resetting}, which is a simple but surprisingly effective algorithm for solving random SAT instances that are meant to be hard. In particular, an evaluation on the Random Track of the 2017 and 2018 SAT competitions shows that a basic prototype of this simple idea already ranks at second place in both years. We share this observation in the hope that it helps the SAT community better understand the hardness of random instances used in competitions and inspire other interesting ideas on SAT solving. 
\end{abstract}

\section{Introduction}

Boolean satisfiability problem (SAT for short)  is the first well-known NP-complete problem~\citep{Cook:1971}.
Though SAT is theoretically challenging, tremendous progress has been made in practice. 
On one hand, efficient algorithms have been developed and gradually refined over time.
Two prominent approaches are \textit{tree search}~\citep{Davis:1962}, in particular conflict-driven clause learning (CDCL) \citep{CDCL}, and \textit{local search}~\citep{GSAT}.
On the other hand,
a challenging and properly designed suite of benchmarks could greatly help research, since novel ideas can be quickly evaluated and compared against each other. 
SAT competitions %~\citet{sat-comp} 
provide such an ideal research platform. 
The central problem of setting up a benchmark suite is to design interesting and challenging instances~\citep{barthel2001hiding} so that different ideas are encouraged. 
Indeed, the two dominant approaches are routinely winners of SAT competitions. 
In particular, top ranking solvers of the Main Track (composed of SAT instances from applications) are dominated by CDCL-based solvers, while local search solvers perform well on the Random Track (composed of randomly generated instances targeting weaknesses of the existing solvers \citep{barthel2001hiding, qhid, komb}). % \yujia{some citations here would be great}.

A fundamental component of CDCL solvers is the unit propagation procedure, which starts from a partial assignment of the variables, and then iteratively uses resolution rules to infer the assignments of other variables, or derive a conflict.  For example, given a clause $(x_1 \lor x_2 \lor \lnot x_3)$, and $x_2=$ False, $x_3=$ True, in order to satisfy this clause, $x_1$ must be True.  Now that we have inferred the value of $x_1$ given $x_2$ and $x_3$, the value of $x_1$ may be used to infer the assignment of other variables in other clauses.

% Unit propagation is the very basic component of any tree search approach, which is arguably more fundamental than the CDCL heuristic. 
%
% However, this powerful unit propagation procedure has so far not been utilized by any good local search solvers, which takes an initial assignment of all variables and then apply local moves to tweak the assignment and make progress towards a fully satisfying assignment.  
% But it has not been explored in any local search approach that is known to be good at solving random instances. 
% One obvious reason for this might be that local search solvers work with complete rather than partial assignments, while unit propagation is only applicable to partial assignments.
% The reason seems to be quite obvious that unit propagation is only useful when variable assignments are partially available, while local search approach always starts with a full assignment. 
%

In this short note, we present a simple variant of unit propagation that works on full assignments, 
which, combined with a periodic resetting schedule, forms the entire SAT solving algorithm.
% No heuristics like variable selection, polarity selection or clause learning are needed.  [Yujia: We do have a heuristic for choosing variable ordering.]
Other than randomly initializing the full assignments at the beginning of the SAT solving process, the algorithm is completely deterministic. 
% We then propose a very simple local search algorithm using this modified unit propagation as the driving force for the search.
% However, our investigation indicates that this is not case. 
%
% We find that a simple variant of unit propagation can replace random flips and be used as the only driving force during local search.
%
% One may expect that such a simple idea could quickly reach a local optimum and get stuck. 
%
Surprisingly, evaluation on the Random Track of the 2017 and 2018 SAT competitions shows that this seemingly simple approach alone can rank at the second place in both competitions, outperforming much more sophisticated approaches.

% However, our study shows that, with periodic resetting, its performance is actually quite close to the performance of the SAT competition winner at least on random track. 

% \begin{itemize}
%     \item Boolean satisfiability in general
%     \item CDCL
%     \item local search
%     \item hard instance generation
%     \item iterative unit propagation
% \end{itemize}

\section{Approach}

\subsection{Background}
\textbf{Conjunctive normal form}.
The conventional representation of a SAT problem is the conjunctive normal form (CNF), which consists of a conjunction of \emph{clauses}. Each clause is a disjunction of \emph{literals}, and a literal is either a Boolean \emph{variable} or its negation. 
For example, $(x_1 \vee x_2 \vee \neg x_3 ) \wedge (x_3 \vee \neg x_1 \vee x_4)  $
is a SAT problem (or instance) with two clauses, each consisting of three literals. 

\textbf{Notation}.
We will use $\phi$ to denote a SAT instance, $V=\{x_1, ..., x_n\}$ to denote the corresponding set of Boolean variables, 
and $\alpha$ to denote a full assignment, with $\alpha_i$ denoting the assigned truth value for variable $x_i$, and $\phi(\alpha)$ the truth value of assignment $\alpha$ for instance $\phi$.
% and $S$ to denote a candidate solution, which may or may not be satisfiable, e.g. $S = \{x_1=\bot, x_2=\bot, x_3=\top, x_4=\top\}$, where $\top (\bot)$ means $True (False)$. 

\textbf{Unit propagation}. Assuming there exists at least one satisfiable assignment for a given partial assignment, unit propagation (or Boolean constraint propagation) is a technique that forces 
the clause with a single literal (a \emph{unit clause}) to be $True$ by making that literal being $True$.
This can be generalized to clauses with multiple literals in the case where all but one of the literals are known to be $False$.  To apply unit propagation in this case, we need to (1) simplify the instance by removing all the literals that evaluate to $False$ given the current partial assignment, and (2) set the variables in unit clauses according to their sign in the clauses.
For instance, suppose we have a partial assignment $\{x_1=False, x_2=False\}$ for our above example, the first clause $(x_1 \vee x_2 \vee \neg x_3 )$  can be simplified to $(\neg x_3)$. Unit propagation will force $\neg x_3$ to be $True$. Thus, $x_3$ should be $False$, and in a similar way with the second clause, unit propagation will force $x_4$ to be $True$.
A more formal description of unit propagation is given in Algorithm~\ref{up-alg}.

\begin{algorithm}[H]
\KwIn{SAT instance $\phi$, partial assignment $\alpha_{partial} $}
\KwOut{New partial assignment $\alpha_{partial}'$}
\SetAlgoLined

\DontPrintSemicolon
 $\alpha_{partial}' \leftarrow \alpha_{partial} $ \;
 \Repeat{no updates on $\alpha_{partial}'$}{
   $\phi' \leftarrow $ simplify $\phi$ according to $\alpha_{partial}'$ \;
   \lIf{there exists any unit clause $c \in \phi'$ }{
    $\alpha_{partial}' \leftarrow \alpha_{partial}' \cup \{var(c) = sign(c) \}$ 
   }
  } 
 \KwRet{$\alpha_{partial}'$}
 \caption{Unit propagation}
 \label{up-alg}
\end{algorithm}

% \vspace{-10em}
\subsection{Prioritized Unit Propagation}
We now introduce \textit{prioritized unit propagation}, a simple extension of the standard unit propagation that can be used as an improvement operator for full assignments, rather than partial assignments.  Our algorithm (Algorithm~\ref{pup-alg}) constructs a new full assignment from the current assignment by first picking an ordering of the variables (line 1-3, hence the name `prioritized'). It initializes the new assignment to be empty (line 4), assigns one variable at a time to its current value following the ordering (line 8), and after the assignment of each variable, runs unit propagation on the partial assignment to set more variables (line 9). This is done iteratively until all the variables are set.

The particular variable ordering we use is based on the variance of variable assignments across invocations of the prioritized unit propagation procedure.  Our algorithm 
% Algorithm~\ref{pup-alg} illustrates our key idea that maintains priorities of variables, according to which standard unit propagation is applied iteratively. 
% In particular, it 
maintains an exponential moving average (EMA) for each variable assignment, where a $True$ value is interpreted as 1, and a $False$ value is interpreted as 0, and then we compute the variance of the assignment based on the EMA (line 2), which determines the variable ordering.  We prioritize variables with larger variances. This has the effect of first assigning values to variables that change the most across iterations, with the more stable variables being more likely to be assigned values by unit propagation. We speculate that such a prioritization results in more exploration and less likelihood of getting stuck in local optima. We have also tried prioritizing variables in increasing order of variance, but the results are worse.

% of variables (line 3)  that will be visited in the subsequent loop (line 5 - 7). 
% %
% The loop then invokes the \textit{standard} unit propagation by clamping the first $n$ variable assignments of the current candidate solution as a \textit{partial} solution.
% We note that this loop can be implemented in a much more efficient way by leveraging dynamic programming trick during unit propagation, while the current presentation is preferred due to its simplicity.

% The nice property of prioritized unit propagation is that, 
% once it reaches a satisfiable assignment, the algorithm converges and makes no more updates.
% Thus, it will never destroy any satisfiable assignment. 

\begin{algorithm}[t]
\KwIn{SAT instance $\phi$, assignment $\alpha$, exponential moving average of assignment values $ema$}
\KwOut{Updated assignment $\alpha'$ and $ema$}
\SetAlgoLined

\SetKwFunction{FuncUP}{UnitProp}
\SetKwFunction{FuncUpdatePriority}{UpdatePriority}
\SetKwProg{Fn}{Function}{:}{}
\DontPrintSemicolon
% sort variables according variances \;

 $ema \leftarrow ema * \rho + \alpha * (1 - \rho) $ \tcp*{Here $\alpha$ is interpreted as a vector}
 $variance \leftarrow ema * (1 - ema) $ \;
 $R \leftarrow $ rank variable by variances (high to low) \; 
%  $\alpha' \leftarrow \alpha $ \;
%  \For{$n\gets 1$ \KwTo $|V|$}{
%   $\alpha_{partial} \leftarrow $ assignments of first $n$ variables of $\alpha'$ according to $R$ \;
%   $\alpha' \leftarrow$  \FuncUP{$\phi, \alpha_{partial}$}
%  }
%  \KwRet{$\alpha'$}
 $\alpha' \leftarrow \{\} $ \;
 \For{$n\gets 1$ \KwTo $|V|$}{
  $i \leftarrow $ the $n$-th variable in ordering $R$ \;
  \If{$\alpha'$ does not include assignment for $x_i$}{
    Add assignment $\{x_i = \alpha_i\}$ to $\alpha'$ \;
    $\alpha' \leftarrow$ \FuncUP{$\phi, \alpha'$} \;
  }
 }

 \KwRet{$\alpha', ema$}

 \caption{Prioritized Unit Propagation}
 \label{pup-alg}
\end{algorithm}

\subsection{Periodic Resetting}
Prioritized unit propagation can make rapid progress towards satisfying more clauses and may eventually converge on a satisfying assignment.
But it could get stuck on an unsatisfying assignment as well. 
To mitigate this issue, we introduce another simple technique, namely \textit{periodic resetting}, inspired by the restarting mechanism in modern SAT solvers. 
Unlike restarting, peridoic resetting restores the current assignment to the best assignment so far (i.e., satisfying the most number of clauses).
Algorithm~\ref{reset-alg} summarizes our main SAT solving algorithm, \emph{Prioritized Unit Propagation with PEriodic Resetting} (PUPPER), which combines prioritized unit propagation and periodic resetting. It starts with a random assignment (line 2), keeps calling prioritized unit propagation (line 6), and resets the assignment according to the given frequency (line 7). 
It terminates when either the maximum number of iterations is reached or a satisfying assignment is found. 

\begin{algorithm}[H]
\KwIn{SAT instance $\phi$, maximum number of iterations $N$}
\KwOut{Assignment $\alpha$ (which may or may not be satisfying)}
\SetAlgoLined

\SetKwFunction{FuncPUP}{PrioritizedUnitProp}
\SetKwFunction{FuncRI}{RandomInit}
\SetKwFunction{FuncSol}{Solve}
\SetKwProg{Fn}{Function}{:}{}
\DontPrintSemicolon

 $ n \leftarrow $ 0\;
 $ \alpha \leftarrow $ randomly assign $True$ or $False$ value to each variable \;
 $ \alpha_{best} \leftarrow \alpha$ \;
 $ema \leftarrow \alpha$ \;

 \While{++n $<$ N and $\phi(\alpha)$ = $False$ }{
  $ \alpha, ema \leftarrow$ \FuncPUP{$\phi, \alpha, ema$} \;
  \lIf{n \% frequency = 0}{
    $\alpha \leftarrow \alpha_{best}$
  }
  \lIf{$\alpha$ satisfies more clauses than $\alpha_{best}$}{
    $\alpha_{best} \leftarrow \alpha$}
  \lIf{$\phi(\alpha_{best}) = True$}{
    \KwRet{$\alpha_{best}$}}     
 }
 \KwRet{$\alpha_{best}$}

 \caption{Prioritized Unit Propagation with PEriodic Resetting (PUPPER)}
 \label{reset-alg}
\end{algorithm}

\subsection{Parallelization}
A simple way to parallelize our algorithm is to run multiple independent copies, each with a different random initialization. Running time to find a satisfying assignment is expected to improve by running multiple copies in parallel and stopping once any copy finds one.
We show another very interesting observation in the evaluation below, that even with a \textit{single} thread, running multiple copies still significantly improves the performance. 

% Of course, this is unsurprising, as it is a property that any local search approach, or more generally undeterministic approach, exhibits. (let's never say whether our algorithm is local search or not)

\section{Evaluation}

% SAT competition hardware: Intel(R) Xeon(R) CPU E5-2609 0 @ 2.40GHz (2393 MHZ)
\subsection{Implementation and Evaluation Setup}
We implement a prototype of \toolname{} in C++ using the standard STL containers without any tuning for high performance. 
We note that modern SAT solvers greatly benefit from efficient data structure manipulations, e.g. the 2-watched literal trick~\citep{chaff}, which we could benefit from as well, but didn't implement it in this prototype as these are not our focus.

We evaluate \toolname{} on the Random Track instances from the 2017 and 2018 SAT competitions.  For the evaluation, the maximum number of iterations is set to 1 million, and resetting happens every 5 iterations.
The SAT competition uses Intel(R) Xeon(R) CPU E5-2609 with 2.40 GHz frequency, 264 GB memory to evaluate participant solvers. 
We do not have hardware with the same configuration, thus instead, we used our own hardware (Intel Skylake Xeon CPUs 2.0 GHz) for the evaluation.
All evaluations are done with a single thread, so our solver only uses a single core. Also, our evaluation was done on a shared cluster, so the timing of our evaluation could be negatively affected, i.e., the actual running time when evaluated on a dedicated machine could be even smaller than the numbers reported here.

\subsection{Main Results}
Table~\ref{eval:table} summarizes our evaluation results. 
The second and third columns list the number of solved instances and the average, median, and maximum running time for each solver (counting only the solved instances) on Random Track benchmarks from the 2017 and 2018 SAT competitions, respectively. 
The second row shows the performance of the winner solver and the second place solver. Note that the winner as well as the runner-up refer to different solvers in different years.  
The third row shows the performance of \toolname{} with different number of copies (running in a single thread). 
The last two rows are two ablation studies: 1) ``no priorities" row shows the results of running with 32 copies but without maintaining priorities, and instead, a random order is used; 2) ``no periodic resetting" row shows the results of running with 32 copies and prioritized unit propagation but without periodic resetting.

The interesting observation is that, with more copies, \toolname{} not only solves more instances but also solves them faster. 
However, this gain could saturate with enough number of copies, since only a single thread is used and the amount of time spent on a particular copy will decrease with more copies.
As we can see, such an effect happens when we increase the number of copies from 32 to 64.
Note that such an issue can be easily addressed by allocating a thread for each copy, and doing so will immediately give us a linear speedup. 
In addition, the two ablation studies suggest that both priorities and periodic resetting are crucial parts of \toolname{}. More specifically, the performance of \toolname{} is affected relatively more by priorities. 

In general, Table~\ref{eval:table} indicates that 
\toolname{} already outperforms the second place solver, which is surprising given the simplicity of our idea and
the fact that top ranking participants are usually rich combinations of many heuristics and optimizations that are specialized for different tracks of benchmark suites. 
For example, the winner of the SAT competition 2018 Random Track combines a local search solver and a CDCL solver. 
It is clear that \toolname{} on its own is not yet as good as the winner, but the gap between them is quite small.

\begin{table}
{\small
    \centering
    \caption{Performance on 2017 and 2018 SAT competition Random Track}
    \label{multiprogram}
    \hspace{-1em}
    \begin{tabular}{|c | r  r | r  r| }
    \hline
    \multirow{3}{*}{Solvers}
        & \multicolumn{2}{c|}{2017 Random Track} & \multicolumn{2}{c|}{2018 Random Track} \\
        % & \#solved 
        % & average / median / max time (seconds)
        &  \begin{tabular}{@{}c@{}} \#solved  \\  instance \end{tabular}
        &  \begin{tabular}{@{}c@{}}average / median / maximum  \\ time (seconds)\end{tabular}
        % & \#solved 
        % & average / median / max time (seconds)
        &  \begin{tabular}{@{}c@{}}\# solved  \\  instance \end{tabular}
        &  \begin{tabular}{@{}c@{}}average / median / maximum  \\ time (seconds)\end{tabular}
        \\
    
        \hline
    
        Winner    & 124 & 276 / 11 / 3631  & 188 &  93 / 0.12 / 502 \\
        Runner-up & 113 & 220 / 50 / 4905     & 165 & 9.4 / 1.1 / 413 \\

        \hline
        % with single thread
        \toolname{} (\#copies=1)  & 101 & 69 / 8.94 / 693  &  157 &  7.0  /  0.43\,  /  320 \\
        \toolname{} (\#copies=2)  & 111 & 70 / 7.58 / 956  &  165 &  4.0  /  0.46\,  /  152 \\
        \toolname{} (\#copies=4)  & 118 & 84 / 7.78 / 1354 &  165 &  4.1  /  0.43\,  /  386 \\
        \toolname{} (\#copies=8)  & 118 & 96 / 7.43 / 4458 &  165 &  1.7  /  0.35\,  / \, \ 60 \\
        \toolname{} (\#copies=16) & 119 & 53 / 8.44 / 852 &  165 &  1.8  /  0.42\,  / \, \ 90 \\
        \toolname{} (\#copies=32) & 120 & 65 / 6.24 / 3916 &  165 &  1.6  /  0.51\,  / \, \ 19 \\
        \toolname{} (\#copies=64) & 120 & 84 / 7.24 / 5858  &  165 &  1.8  /  0.71\,  / \, \ 42 \\

        \hline
        
        % \begin{tabular}{@{}c@{}} \toolname{}(\#copies=32)  \\  w/ random order \end{tabular} 
        no priorities & 49 & 3.39 / 1.4 / 23 & 119 & 713 / 0.86 / 12011 \\
        \hline
        %\begin{tabular}{@{}c@{}} \toolname{}(\#copies=32)  \\  w/o periodic resetting \end{tabular} 
        no periodic resetting & 101 & 1276 / 176 / 13670 & 165 & 62 / 1.60 / 1809\\
        
        \hline
        % 16 thread
        %\begin{tabular}{@{}c@{}} \#thread=1, \#copies=1 \end{tabular}
        % IUP & 119 & 28 / 0.36 / 876  & 165 &  max \\
        % \hline

        % \cline{2-9}
        %  & \multicolumn{8}{|c|}{Sets}\\
        % \cline{2-9}
        %  & 1 & 2 & 3 & 4 & 5 & 6 & 7 & 8\\
        % \hline
        % \multicolumn{1}{|c|}{astar} & & * &  & * &  &  & * &\\
        % \hline
    \end{tabular}
    \label{eval:table}
    }
\end{table}

\section{Discussion}

% We propose prioritized unit propagation with periodic resetting, which is simple and effective  
% but has not been explored in local search approach.
% Our evaluation on the recent SAT competition random track benchmark suite indicates this simple idea is fairly encouraging.
The fact that prioritized unit propagation with periodic resetting works surprisingly well on the recent SAT competition Random Track benchmark suite suggests a few important implications on SAT solving as well as benchmark setup. 

First of all, we believe variations of unit propagation deserves further exploration in both tree search and local search approaches. Indeed, there are already a few studies on this. Two closely related ones are UnitWalk~\citep{UnitWalk} and EagleUP~\citep{EagleUP}. Both work use unit propagation as a way of improving local search, but the priorities and periodic resetting are not considered, which as we show are very important as well. 
Nevertheless, our way of maintaining priorities and resetting periodically is fairly preliminary, which could be further improved. 

On the other hand, we observe all benchmarks solved by \toolname{} are generated according to three algorithms published in the literature \citep{barthel2001hiding, qhid, komb}.
This observation suggests that new challenging benchmark suites are perhaps needed for future SAT competitions, and our idea could be a strong baseline for evaluating new benchmark generation algorithms.
% help  better understand the hardness of the recent and design stronger benchmark suite. 
% We develop a basic prototype and evaluate it on the Random track of SAT competition 2017 and 2018.  

\bibliography{iclr2020_conference}
\bibliographystyle{iclr2020_conference}
% \bibliographystyle{ieeetr}

% \appendix
% \section{Appendix}

\end{document}